\documentclass[letterpaper, 10 pt, conference]{ieeeconf}  

\IEEEoverridecommandlockouts                              

\usepackage{graphics} 
\usepackage{epsfig} 
\usepackage{mathptmx} 
\usepackage{times} 
\usepackage{amsmath} 
\usepackage{amssymb}  
\usepackage{cite}
\usepackage{algorithmic}
\usepackage{graphicx}
\usepackage{textcomp}
\usepackage{xcolor}
\usepackage{multirow}
\usepackage{hyperref}

\title{SpikingRTNH: Spiking Neural Network for 4D Radar Object Detection}

\author{Dong-Hee Paek and Seung-Hyun Kong$^*$
\thanks{*Corresponding author}
\thanks{D.-H. Paek and S.-H. Kong are with the CCS Graduate School of Mobility, KAIST, Daejeon 34051, Republic of Korea. \texttt{\{donghee.paek, skong\}@kaist.ac.kr}}
}

\begin{document}

\maketitle
\thispagestyle{empty}
\pagestyle{empty}

\maketitle

\begin{abstract}
Recently, 4D Radar has emerged as a crucial sensor for 3D object detection in autonomous vehicles, offering both stable perception in adverse weather and high-density point clouds for object shape recognition. However, processing such high-density data demands substantial computational resources and energy consumption. We propose SpikingRTNH, the first spiking neural network (SNN) for 3D object detection using 4D Radar data. By replacing conventional ReLU activation functions with leaky integrate-and-fire (LIF) spiking neurons, SpikingRTNH achieves significant energy efficiency gains. Furthermore, inspired by human cognitive processes, we introduce biological top-down inference (BTI), which processes point clouds sequentially from higher to lower densities. This approach effectively utilizes points with lower noise and higher importance for detection. Experiments on K-Radar dataset demonstrate that SpikingRTNH with BTI significantly reduces energy consumption by 78\% while achieving comparable detection performance to its ANN counterpart (51.1\% AP\textsubscript{3D}, 57.0\% AP\textsubscript{BEV}). These results establish the viability of SNNs for energy-efficient 4D Radar-based object detection in autonomous driving systems. All codes are available at \url{https://github.com/kaist-avelab/k-radar}.
\end{abstract}


\section{Introduction}

The advancement of 4D Radar technology has revolutionized autonomous driving perception \cite{radar_survey}. 4D Radar precisely measures the position and movement of surrounding objects through range, azimuth angle, elevation angle, and Doppler frequency. Unlike camera and LiDAR, 4D Radar exhibits superior stability under adverse weather conditions such as rain and snow \cite{kradar}. Moreover, compared to conventional 3D Radar \cite{automotive_review} that mainly focus on object presence recognition, 4D Radar provides crucial height information through elevation angle measurements, enabling detailed 3D shape recognition of objects. Consequently, 4D Radar has become a key sensor for robust 3D object detection \cite{enhanced_kradar,rtnhp,dual}.

\begin{figure}[tb!]
    \centering
    \includegraphics[width=0.48\textwidth]{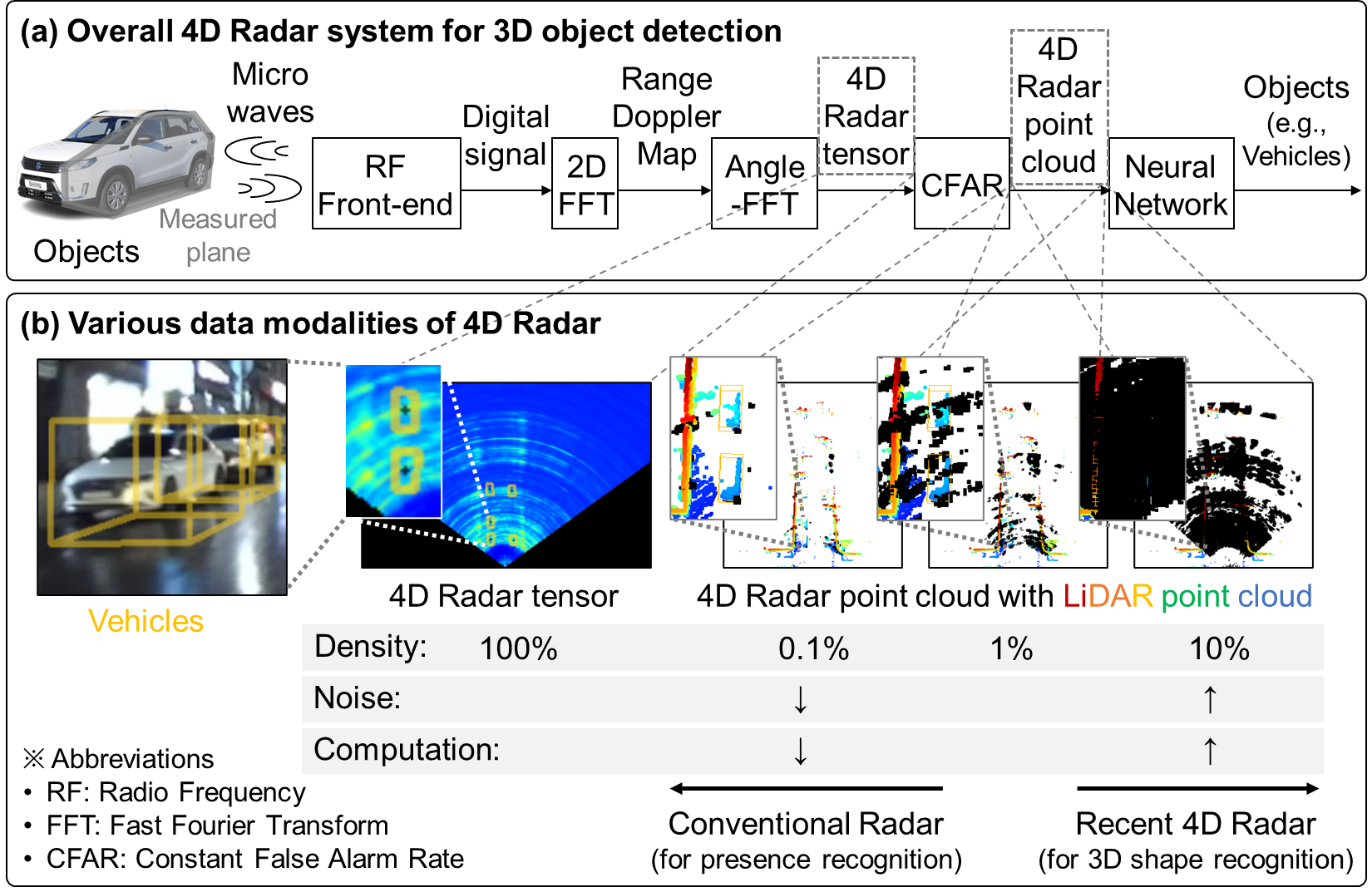}
    \caption{4D Radar system and data representation: (a) Overview of the 4D Radar-based 3D object detection system. Radio waves reflected from target objects (e.g., vehicles) are converted into digital signals in RF front-end. A 2D FFT then generates a Range-Doppler Map, and an Angle-FFT extracts the 4D Radar tensor. Next, CFAR filtering produces the 4D Radar point cloud, which is finally processed by a neural network to detect objects. (b) Various data modalities of 4D Radar shown with different point cloud densities. From left to right: a camera image (with a yellow 3D bounding box indicating a vehicle), a raw 4D Radar tensor (showing all measurements), and 4D Radar point clouds at different densities (0.1\%, 1\%, 10\%). A LiDAR point cloud is overlaid for reference. The bottom row compares key characteristics (Density, Noise, Computation) between conventional Radar and recent 4D Radar \cite{radar_tutorial}. While conventional Radar uses approximately 0.1\% of data points for presence detection (resulting in lower noise and computational cost), modern 4D Radar leverages around 10\% density for detailed 3D shape recognition (leading to higher noise but richer shape information).}
    \label{fig:radar_system}
\end{figure}

As illustrated in Fig.~\ref{fig:radar_system}, modern 4D Radars generate rich sensor data in two forms: 4D Radar tensor (4DRT) and 4D Radar point cloud (4DRPC). 4DRPC with 10\% density provides approximately 100 times more measurement points than conventional Radar systems, enabling high-precision object shape recognition. Radar Tensor Network with Height (RTNH) \cite{kradar} pioneered 3D object detection using 4D Radar by processing the top 10\% power signals from the 4DRT. Subsequent research has shown that utilizing high-density points (1-5\% of measurements) while filtering sidelobe noise can further enhance detection performance \cite{rtnhp}.

However, 4D Radar-based 3D object detection faces a significant challenge: the high energy consumption required to process dense point cloud data. For instance, RTNH \cite{kradar} requires approximately 156G multiply-accumulate (MAC) operations, consuming 7.16 J of energy per frame during inference. This substantial energy requirement can impact the overall efficiency and sustainability of autonomous driving systems.

To address this challenge, brain-inspired spiking neural networks (SNNs) have gained attention in autonomous driving applications \cite{snn_survey,direct_snn_survey,dl_in_snn}. SNNs employ spike-event-based computing, performing simple accumulation (AC) operations that substantially reduce energy consumption compared to conventional artificial neural networks (ANNs) \cite{snn_efficiency}. The energy efficiency of SNNs has been demonstrated across various autonomous driving tasks. For example, \cite{snn_end_to_end} implemented an end-to-end autonomous driving pipeline using SNNs, achieving up to 98.7\% energy reduction in the planning module compared to state-of-the-art ANN models \cite{planning_sota}. Similarly, \cite{lane_snn} developed an SNN for lane detection using event-camera images that requires only 1W of power, while \cite{da_snn} demonstrated that an SNN for drivable-area detection using LiDAR point clouds reduced energy consumption by up to 98.8\% while maintaining comparable accuracy.

Building on these advances, we present \emph{SpikingRTNH}, the first SNN architecture specifically designed for 4D Radar-based 3D object detection. Our approach offers significantly higher energy efficiency compared to its ANN counterpart (RTNH \cite{kradar}). Inspired by \cite{snn_direct_training}, we replace ANN neurons (ReLU activation) with leaky integrate-and-fire (LIF) spiking neurons to enable training on standard CPU/GPU hardware. Additionally, drawing from human top-down cognitive processes \cite{top_down_perception,top_down_perception_infant}, we propose a \emph{biological top-down inference} (BTI) method that processes point clouds from high density to low density, leveraging the fact that lower-density 4DRPC contains fewer points with reduced noise.

In summary, our contributions are as follow,
\begin{itemize}
    \item We propose SpikingRTNH, the first SNN architecture for 4D Radar-based 3D object detection, demonstrating the viability of energy-efficient spike-based computing.
    \item We introduce biological top-down inference (BTI), a novel approach inspired by human cognitive processes that achieves 51.1\% AP\textsubscript{3D}, 57.0\% AP\textsubscript{BEV} comparable to its ANN counterpart.
    \item We achieve a 78\% reduction in energy consumption, significantly advancing the efficiency of 4D Radar-based perception systems.
\end{itemize}

The remainder of this paper is organized as follows: Section II reviews related work in SNNs and 4D Radar-based 3D object detection. Section III details the architecture of SpikingRTNH and the proposed BTI approach. Section IV presents experimental results validating our method's performance, and Section V concludes the paper.

\section{Related Work}

\subsection{Spiking Neural Network}

Spiking neural networks (SNNs) aim to replicate biological neuron behavior in artificial systems. Unlike ANN neurons that process continuous values in a temporally static manner, SNN neurons communicate through discrete spikes and incorporate temporal dynamics \cite{snn_survey}. While various neuron models exist, the leaky integrate-and-fire (LIF) model has gained widespread adoption due to its biological plausibility and implementation efficiency \cite{lif}. This spike-based computation eliminates multiplication operations from the conventional multiply-accumulate (MAC) operations in ANN layers, utilizing accumulation (AC) operations that significantly improve energy efficiency \cite{snn_efficiency,snn_first_layer}. In 45\,nm CMOS technology, a single MAC operation in an ANN consumes approximately 4.6\,pJ, whereas an AC operation in an SNN requires only 0.9\,pJ \cite{energy_comparison}. Furthermore, SNNs activate neurons only when spike events occur, unlike ANNs on GPUs where all neurons remain active \cite{snn_survey}. This characteristic is particularly advantageous for sparse data like point clouds, where most neurons remain inactive, maximizing computational efficiency.

\subsection{Object Detection Networks for 4D Radar}

The development of large-scale datasets spanning diverse conditions has been crucial for advancing sensor-based object detection networks \cite{kitti,nuscenes}. Several 4D Radar datasets have recently emerged, catalyzing research in 4D Radar-based object detection. Astyx \cite{astyx} pioneered open-access 4D Radar point clouds but contains only 0.5K frames, limiting its utility for training high-performance networks. VoD \cite{vod} and TJ4DRadSet \cite{tj4d} provide 8.7K and 7.8K frames respectively, but primarily focus on urban driving scenarios under normal weather conditions. K-Radar \cite{kradar} offers 35K frames of raw 4D Radar measurements (4DRT), collected across diverse road environments and weather conditions. Dual Radar \cite{dual} employs two Radar sensors with both dense and sparse outputs to demonstrate the importance of high-density data in 3D object detection.

These datasets have enabled the development of various 4D Radar object detection networks. Early works \cite{rpfa,vod,tj4d} adapted PointPillars architecture \cite{pointpillars} to demonstrate the feasibility of 3D object detection using 4D Radar. RTNH \cite{kradar} achieved robust 3D object detection on K-Radar by processing the top 10\% power signals from 4DRT. Recent improvements include filtering sidelobe noise and incorporating vertical information through attention mechanisms \cite{rtnhp}, as well as enhancing performance by integrating Doppler measurements in data accumulation or encoder modules \cite{multiframe_4d,rcfusion}. However, these methods rely on GPU-based ANN models, resulting in high energy consumption. Our work addresses this limitation by introducing SpikingRTNH, the first energy-efficient SNN architecture for 4D Radar-based 3D object detection.

\section{Methodology}

\begin{figure*}[tb!]
    \centering
    \includegraphics[width=0.98\textwidth]{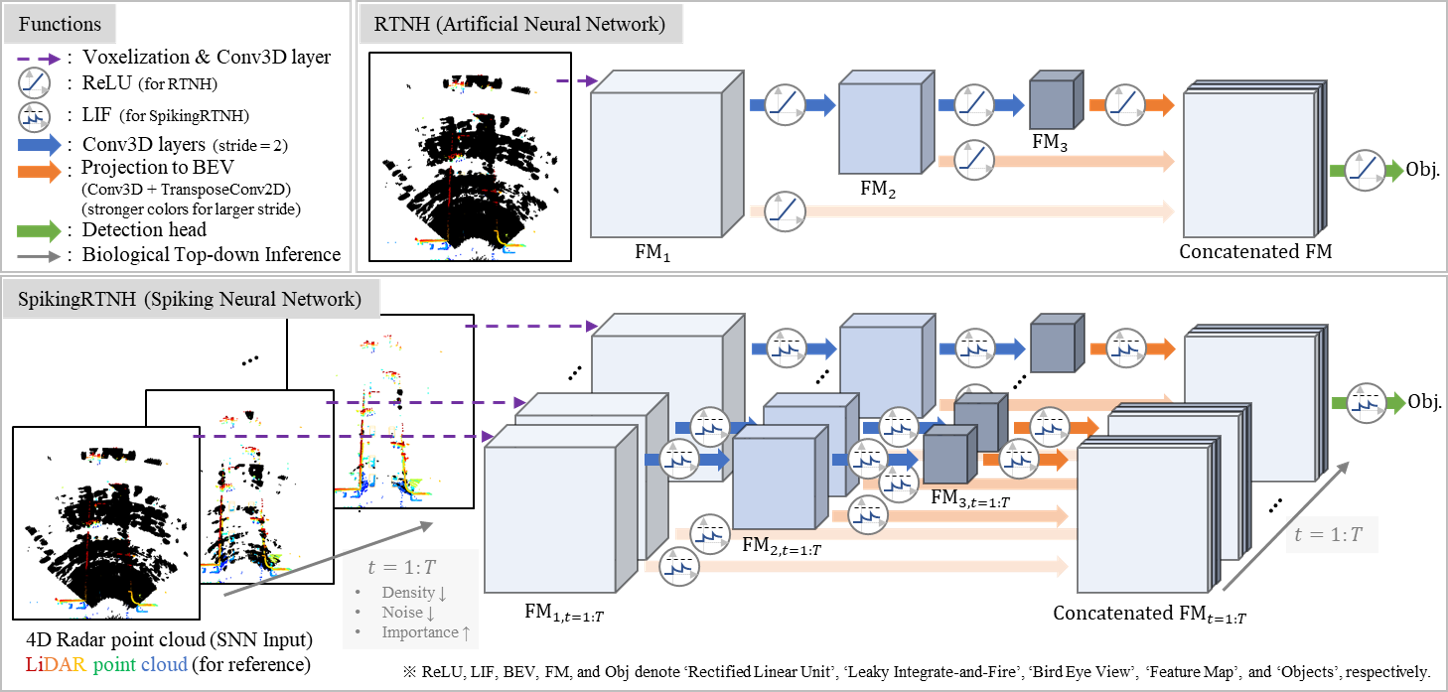}
    \caption{Network architecture comparison between RTNH and SpikingRTNH. Top: RTNH processes a single high-density 4D Radar point cloud using ReLU activation functions, generating three stages of feature maps (FM\textsubscript{1,2,3}). The feature maps are projected to bird's-eye-view (BEV) and concatenated for final object detection. Bottom: SpikingRTNH sequentially processes point clouds from higher to lower densities ($t$=1:$T$) by replacing ReLU with leaky integrate-and-fire (LIF) neurons. This biological top-down inference (BTI) approach leverages the observation that lower-density point clouds contain fewer noisy points while retaining the most critical points for object detection. LiDAR point cloud is included for reference only and is not used in the network processing.}
    \label{fig:net_structure}
\end{figure*}

This section details the proposed SpikingRTNH architecture and our inference strategy grounded in biological cognitive mechanisms. In Section~\ref{subsec:spiking_rtnh}, we describe the conversion of RTNH to an SNN architecture and present our network structure. In Section~\ref{subsec:bio_mechanism}, we introduce our biological top-down inference (BTI) method inspired by human cognitive processes.

\subsection{RTNH to SpikingRTNH}
\label{subsec:spiking_rtnh}

\noindent\textbf{RTNH.}
As shown in Fig.~\ref{fig:net_structure}, Radar Tensor Network with Height (RTNH) \cite{kradar} processes high-density (approximately 10\%) 4D Radar point clouds (4DRPC) extracted from 4D Radar tensors (4DRT). To formally define our approach, we first represent the 4DRPC as:
\begin{equation}
    \mathcal{P} = \left\{ \mathbf{p}_i \mid \mathbf{p}_i \in \mathbb{R}^5, i = 1, 2, \dots, N \right\},
\end{equation}
where each point $\mathbf{p}_i = (x_i, y_i, z_i, {pw}_i, {dop}_i)$ represents 3D coordinates, power, and Doppler measurements, and $N$ denotes the total number of points. RTNH first voxelizes these points and applies a 3D convolution layer to generate the first-stage feature map:
\begin{equation}
\label{eq:fm}
    \mathrm{\mathbf{FM}_1} \in \mathbb{R}^{C_{\mathrm{1}} \times Z_{\mathrm{1}} \times Y_{\mathrm{1}} \times X_{\mathrm{1}}},
\end{equation}
where $C_1, Z_1, Y_1, X_1$ represent the channel and spatial dimensions of $\mathrm{\mathbf{FM}}_1$, respectively.

RTNH processes this feature map through three stages of 3D convolutions, batch normalization, and ReLU activations (see Fig.~\ref{fig:net_structure}), generating three feature maps ($\mathrm{\mathbf{FM}_1}, \mathrm{\mathbf{FM}_2}, \mathrm{\mathbf{FM}_3}$):
\begin{align}
\label{eq:conv1}
    &\mathrm{\mathbf{FM}}_{j} = f^n_{3D,j-1}(\mathrm{\mathbf{FM}}_{j-1}), \\
\label{eq:conv2}
    &f_{3D}(\cdot) = \mathrm{ReLU}\bigl(\mathrm{Conv3D}(\cdot;\mathbf{K})+\mathbf{b}\bigr),
\end{align}
where $j$ indexes each feature map and $\mathbf{K}$, $\mathbf{b}$ denote the 3D convolution kernel and batchnorm bias, respectively. The last convolution layer has a stride of 2, reducing spatial dimensions by a factor of $2^3=8$; this is repeated three times ($n=3$) to achieve a wide receptive field \cite{vgg}.

To utilize these 3D feature maps collectively, RTNH projects each map onto a bird's-eye-view (BEV) plane using a 3D convolution with kernel size equal to the feature map's $Z$ dimension, followed by a 2D transpose convolution that unifies them into a $Y_1 \times X_1$ sized BEV format. The resulting three BEV feature maps are concatenated \cite{fpn}, and a detection head \cite{yolo} predicts the final objects.

\noindent\textbf{LIF Neuron Model.}
SpikingRTNH adopts the leaky integrate-and-fire (LIF) neuron model \cite{lif}, which offers both biological plausibility and implementation efficiency. An LIF neuron integrates input spikes over time into a membrane potential and fires a spike when the potential exceeds a threshold. The membrane potential $u$ at time $t$ satisfies:
\begin{align}
\label{eq:diff1}
&\tau \frac{du}{dt} = -(u-u_{reset}) + R \cdot I(t), \quad u < V_{th}, \\
\label{eq:diff2}
&\text{fire a spike and } u = u_{reset}, \quad u \geq V_{th},
\end{align}
where $\tau$, $R$, $I$, and $V_{th}$ denote the time constant, resistance, pre-synaptic input, and firing threshold, respectively. Following common practice \cite{snn_direct_training}, $u_{reset}$ is set to 0.

For implementation in deep learning frameworks like PyTorch, we reformulate the continuous equations into discrete time:
\begin{align}
\label{eq:diff3}
&u_k[t+1] = \lambda\bigl(u_k[t]-V_{th}o_k[t]\bigr) + \sum_{l} w_{kl} o_l[t] + b_k,\\
\label{eq:diff4}
&o_k[t] = H\bigl(u_k[t]-V_{th}\bigr),
\end{align}
where $\lambda = 1 - \frac{dt}{\tau}$ is a decay factor (typically set to 0.25 \cite{spiking_pointnet,snn_direct_training}), and $k, l$ index the neurons. $w_{kl}$ and $b_k$ correspond to the kernel and bias in \eqref{eq:conv2}, and $o_k[t]$ represents the spike event at time step $t$ from the $k$-th neuron.

The Heaviside step function $H(\cdot)$ in \eqref{eq:diff4} presents a challenge for gradient-based learning due to its non-differentiability. Following \cite{spiking_pointnet}, we approximate it with a surrogate function:
\begin{equation}
\label{eq:heaviside}
H(u_k[t]-V_{th}) \simeq \frac{1}{2}\tanh\bigl(\beta \,\bigl(u_k[t] - V_{\mathrm{th}}\bigr)\bigr) + \frac{1}{2},
\end{equation}
where $\beta$ (set to 5.0 \cite{spiking_pointnet}) controls the steepness of the approximation.

\noindent\textbf{SpikingRTNH.}
As illustrated in Fig.~\ref{fig:net_structure}, SpikingRTNH maintains the core RTNH pipeline while replacing all ReLU activations with LIF neurons. During training, we employ a single time step ($T=1$) to avoid gradient vanishing/exploding issues that can arise with multiple time steps \cite{spiking_pointnet}. Following \cite{kradar}, we apply focal loss for class imbalance and smooth L1 loss for bounding box regression.

During inference, we leverage SNN temporal dynamics by running multiple time steps ($t$=1:$T$), similar to biological neurons processing information over time. To effectively utilize multi-step inference, we introduce a top-down inference mechanism guided by biological principles.

\subsection{Biological Top-down Inference}
\label{subsec:bio_mechanism}

\noindent\textbf{Cognitive Mechanism.}
Human cognitive mechanism involves both bottom-up and top-down processes \cite{top_down_perception}. Bottom-up processing progresses from basic sensory input to complex features, while top-down processing starts with overall context and moves to specific details. Top-down processing, influenced by prior knowledge, enables controlled attention and reduces cognitive load \cite{top_down_perception_infant}. When searching for critical objects, humans first identify broad features before focusing on specific details \cite{posner}.

\noindent\textbf{Biological Top-down Inference.}
Inspired by this top-down approach—starting from the complete information set before focusing on essential components—we propose biological top-down inference (BTI). This method sequentially processes 4DRPC from higher to lower densities. As shown in Fig.~\ref{fig:radar_system}, lower-density 4DRPC contains fewer total points but also less noise, making it more representative of actual targets.

BTI processes point clouds sequentially from higher to lower density. At discrete time step $t$, the next time step's 4DRPC density is determined by selecting the top $r\%$ of power:
\begin{align}
&{\mathrm{PW}}_{th} = \text{Percentile}_{100-r}\bigl(\{p_{i} \mid i = 1, 2, \dots, N\}\bigr), \\
&\mathcal{P}_{t+1} = \bigl\{ \mathbf{p}_i \,\bigm|\, \mathbf{p}_i \in \mathcal{P}_{t}, \, {pw}_{i} \geq {\mathrm{PW}}_{th} \bigr\}, \quad t=1,2,...,T-1,
\end{align}
where $\mathcal{P}_{t}$ represents the 4DRPC input to SpikingRTNH at time $t$, and $T$ is the total number of time steps. For example, with $r=80\%$, $\mathcal{P}_{3}$ has 64\% of the density of $\mathcal{P}_{1}$. At each time step, the network processes the 4DRPC of the given density to extract feature maps, and the final detection results are obtained from the last time step's feature maps (i.e., $\mathrm{Concatenated FM}_{t=T}$).

BTI offers several advantages: (1) it improves detection accuracy by initially capturing broader context with high-density 4DRPC before refining details with lower-density data, (2) it enhances detection robustness by utilizing noise-reduced lower-density 4DRPC, and (3) it reduces computational load compared to processing 100\% density points over multiple time steps, thereby improving overall efficiency.

\section{Experiments}

In this section, we validate the performance of the proposed SpikingRTNH. First, Section \ref{subsec:setup} presents our experimental setup. Section \ref{subsec:main_results} reports comprehensive comparisons between SpikingRTNH and RTNH. Finally, Section \ref{subsec:ablation} analyzes each component of BTI through ablation studies.

\subsection{Experimental Setup}
\label{subsec:setup}

\begin{figure*}[tb!]
    \centering
    \includegraphics[width=0.84\textwidth]{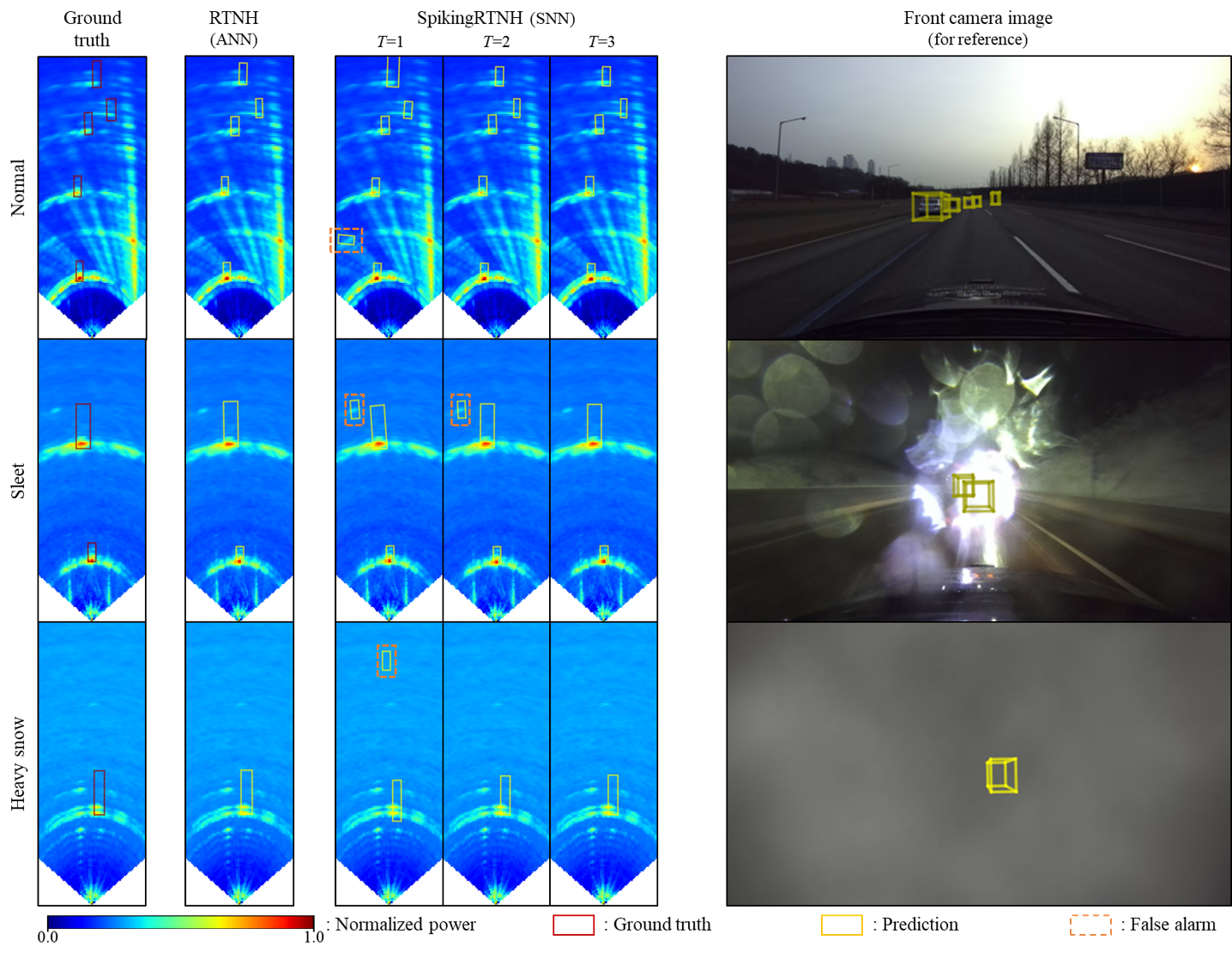}
    \caption{Qualitative comparison of 3D object detection results across different weather conditions (Normal, Sleet, and Heavy snow). The first column shows ground truth bounding boxes (red), while the second column displays RTNH (ANN) predictions (yellow). The next three columns demonstrate SpikingRTNH (SNN) predictions at different time steps ($T$=1,2,3), where solid yellow boxes indicate predictions and dashed orange boxes represent false alarms. The rightmost column shows corresponding front camera images for reference. The heatmap colors represent normalized power values from 0.0 (blue) to 1.0 (red).}
    \label{fig:det_results}
\end{figure*}

\noindent\textbf{Dataset and Metrics.}
We conducted experiments on the K-Radar dataset \cite{kradar}, which provides 35K frames of 4D Radar data collected under various weather conditions (clear, sleet, snow) and road environments (urban, highway). We chose K-Radar as it is currently the only benchmark dataset that provides 4DRT under adverse weather conditions. Following the revised protocol\footnote{Please refer to \href{https://github.com/kaist-avelab/K-Radar/issues/28}{issue \#28} of K-Radar GitHub repository.} of the original paper, we adopted two evaluation metrics for 3D object detection: Average Precision (AP) in 3D (AP\textsubscript{3D}) and BEV (AP\textsubscript{BEV}) for the Sedan class at an IoU threshold of 0.3. Additionally, we used the thop library \cite{thop} to measure computational requirements accurately.

\noindent\textbf{Implementation Details.}
We implemented SpikingRTNH using PyTorch 1.12.0 and trained it on an NVIDIA RTX3090 GPU. We used the AdamW optimizer with a learning rate of 0.001 and weight decay of 0.01. Based on ablation studies, we set the hyperparameters for BTI to $r$ = 80\% and $T$ = 3 time steps.

\subsection{Comparison of SpikingRTNH to RTNH}
\label{subsec:main_results}

\begin{table}[h]
    \caption{Comparison of 3D object detection performance and computational requirements on the K-Radar test set. `-' indicates not applicable.}
    \label{tab:main_results}
    \centering
    \begin{tabular}{l|cc|ccc}
        \hline\hline
        \multirow{2}{*}{Network} & $\text{AP}_{\text{3D}}$ & $\text{AP}_{\text{BEV}}$ & MAC & AC & Energy \\
        & [\%] & [\%] & [/frame] & [/frame] & [J/frame] \\
        \hline
        RTNH \cite{kradar} & 50.7 & 56.5 & 156G & - & 7.16 \\
        \hline
        SpikingRTNH & \multirow{2}{*}{48.1} & \multirow{2}{*}{55.3} & \multirow{2}{*}{2.48G} & \multirow{2}{*}{48.6G} & \multirow{2}{*}{\textbf{0.551}} \\
        (Ours) & & & & & \\
        \hline
        SpikingRTNH & \multirow{2}{*}{\textbf{51.1}} & \multirow{2}{*}{\textbf{57.0}} & \multirow{2}{*}{7.43G} & \multirow{2}{*}{137G} & \multirow{2}{*}{1.58} \\
        + BTI (Ours) & & & & & \\
        \hline\hline
    \end{tabular}
\end{table}

\noindent\textbf{Detection Performance.}
Table \ref{tab:main_results} and Fig. \ref{fig:det_results} present both quantitative and qualitative 3D object detection results on the K-Radar test set. SpikingRTNH ($T=1$) achieves performance comparable to its ANN counterpart RTNH, with only a slight decrease in AP\textsubscript{3D} (48.1\% vs 50.7\%) and AP\textsubscript{BEV} (55.3\% vs 56.5\%). This demonstrates that energy-efficient SNNs can effectively perform 4D Radar object detection. As shown in Fig. \ref{fig:det_results}, SpikingRTNH maintains robust detection capabilities across various weather conditions, including normal, sleet, and heavy snow scenarios.

When applying BTI, SpikingRTNH improves upon the single time step version, achieving gains of 3.0\% and 1.7\% in AP\textsubscript{3D} and AP\textsubscript{BEV} respectively. The qualitative results in Fig. \ref{fig:det_results} clearly demonstrate the progressive improvement in detection quality (i.e., preventing false alarms) as the number of time steps increases from $T=1$ to $T=3$, validating our hypothesis that BTI effectively utilizes lower density points with reduced noise to enhance detection performance.

\noindent\textbf{Energy Analysis.}
In 45nm CMOS technology, a single MAC operation in an ANN consumes approximately 4.6 pJ, while an AC operation in an SNN requires only 0.9 pJ \cite{energy_comparison}. Based on these measurements, RTNH consumes 7.16 J/frame due to its 156G MAC operations. In contrast, SpikingRTNH with BTI requires only 7.43G MAC and 137G AC operations, resulting in significantly lower energy consumption of 1.58 J/frame—a 78\% reduction. The base SpikingRTNH ($T=1$) achieves even greater efficiency with just 0.551 J/frame, representing a 92\% reduction in energy consumption while maintaining reasonable detection performance.

\subsection{Ablation Studies}
\label{subsec:ablation}

\begin{table}[h]
    \caption{Ablation study on BTI parameters. We analyze the impact of time steps ($T$) and density ratio ($r$) on detection performance and computational requirements.}
    \label{tab:ablation_bti}
    \centering
    \begin{tabular}{cc|cc|ccc}
        \hline\hline
        \multirow{2}{*}{$T$} & $r$ & $\text{AP}_{\text{3D}}$ & $\text{AP}_{\text{BEV}}$ & MAC & AC & Energy \\
        & [\%] & [\%] & [\%] & [/frame] & [/frame] & [J/frame] \\
        \hline
        1 & - & 48.1 & 55.3 & 2.48G & 48.6G & \textbf{0.551} \\
        \hline
        2 & 50 & 49.7 & 56.7 & 4.95G & 91.7G & 1.05 \\
        2 & 70 & 49.8 & 56.8 & 4.95G & 93.4G & 1.07 \\
        2 & 80 & 49.8 & 56.7 & 4.95G & 94.2G & 1.08 \\
        2 & 90 & 49.8 & 56.6 & 4.95G & 95.0G & 1.08 \\
        \hline
        3 & 50 & 49.4 & 56.3 & 7.43G & 131G & 1.52 \\
        3 & 70 & 50.0 & 56.8 & 7.43G & 135G & 1.56 \\
        3 & 80 & \textbf{51.1} & \textbf{57.0} & 7.43G & 137G & 1.58 \\
        3 & 90 & 50.9 & 56.9 & 7.43G & 140G & 1.60 \\
        \hline\hline
    \end{tabular}
\end{table}

We conducted ablation studies to analyze the impact of two key parameters of BTI: the number of time steps $T$ and the density ratio $r$. Table \ref{tab:ablation_bti} presents the performance variations across different parameter configurations.

Increasing the number of time steps $T$ generally improves detection performance. The density ratio $r$ shows optimal performance at 80\% at $T=3$, indicating that retaining 80\% of points between consecutive time steps effectively preserves crucial information while progressively reducing noise.

The energy consumption naturally increases with $T$, but even at $T=3$, SpikingRTNH+BTI maintains significantly lower energy requirements compared to RTNH. This demonstrates that our approach successfully balances detection accuracy and energy efficiency.

\section{Conclusion}

This paper presents SpikingRTNH, the first spiking neural network architecture for 4D Radar-based 3D object detection. By replacing conventional ReLU activations with leaky integrate-and-fire (LIF) neurons and introducing biological top-down inference (BTI), we achieve substantial improvements in energy efficiency while maintaining detection performance. Our experiments on the K-Radar dataset demonstrate that SpikingRTNH with BTI reduces energy consumption by 78\% while achieving reasonable performance (51.1\% AP\textsubscript{3D}, 57.0\% AP\textsubscript{BEV}) compared to the conventional ANN approach. The model maintains robust performance across various weather conditions, including challenging scenarios like sleet and heavy snow. This significant reduction in energy consumption while maintaining competitive accuracy demonstrates the practical viability of SNNs for energy-efficient 4D Radar-based object detection in autonomous driving systems.

\section*{Acknowledgment}
This work was supported by the National Research Foundation of Korea (NRF) grant funded by the Korea government (MSIT) (No. 2021R1A2C3008370).

\bibliographystyle{IEEEtran}
\bibliography{egbib}

\begin{thebibliography}{10}
\providecommand{\url}[1]{#1}
\csname url@samestyle\endcsname
\providecommand{\newblock}{\relax}
\providecommand{\bibinfo}[2]{#2}
\providecommand{\BIBentrySTDinterwordspacing}{\spaceskip=0pt\relax}
\providecommand{\BIBentryALTinterwordstretchfactor}{4}
\providecommand{\BIBentryALTinterwordspacing}{\spaceskip=\fontdimen2\font plus
\BIBentryALTinterwordstretchfactor\fontdimen3\font minus \fontdimen4\font\relax}
\providecommand{\BIBforeignlanguage}[2]{{%
\expandafter\ifx\csname l@#1\endcsname\relax
\typeout{** WARNING: IEEEtran.bst: No hyphenation pattern has been}%
\typeout{** loaded for the language `#1'. Using the pattern for}%
\typeout{** the default language instead.}%
\else
\language=\csname l@#1\endcsname
\fi
#2}}
\providecommand{\BIBdecl}{\relax}
\BIBdecl

\bibitem{radar_survey}
L.~Fan, J.~Wang, Y.~Chang, Y.~Li, Y.~Wang, and D.~Cao, ``4d mmwave radar for autonomous driving perception: A comprehensive survey,'' \emph{IEEE Transactions on Intelligent Vehicles}, vol.~9, no.~4, pp. 4606--4620, 2024.

\bibitem{kradar}
D.-H. Paek, S.-H. Kong, and K.~T. Wijaya, ``K-radar: 4d radar object detection for autonomous driving in various weather conditions,'' \emph{Advances in Neural Information Processing Systems}, vol.~35, pp. 3819--3829, 2022.

\bibitem{automotive_review}
S.~M. Patole, M.~Torlak, D.~Wang, and M.~Ali, ``Automotive radars: A review of signal processing techniques,'' \emph{IEEE Signal Processing Magazine}, vol.~34, no.~2, pp. 22--35, 2017.

\bibitem{enhanced_kradar}
D.-H. Paek, S.-H. Kong, and K.~T. Wijaya, ``Enhanced k-radar: Optimal density reduction to improve detection performance and accessibility of 4d radar tensor-based object detection,'' in \emph{2023 IEEE Intelligent Vehicles Symposium (IV)}.\hskip 1em plus 0.5em minus 0.4em\relax IEEE, 2023, pp. 1--6.

\bibitem{rtnhp}
S.-H. Kong, D.-H. Paek, and S.~Lee, ``Rtnh+: Enhanced 4d radar object detection network using two-level preprocessing and vertical encoding,'' \emph{IEEE Transactions on Intelligent Vehicles}, pp. 1--14, 2024.

\bibitem{dual}
X.~Zhang, L.~Wang, J.~Chen, C.~Fang, L.~Yang, Z.~Song, G.~Yang, Y.~Wang, X.~Zhang, and J.~Li, ``Dual radar: A multi-modal dataset with dual 4d radar for autononous driving,'' \emph{arXiv preprint arXiv:2310.07602}, 2023.

\bibitem{radar_tutorial}
\BIBentryALTinterwordspacing
D.-H. Paek and J.~Guan, ``Introduction to {4D} radar: Hardware, {MIMO}, signal processing, dataset, and {AI},'' in \emph{2024 IEEE Intelligent Vehicles Symposium}, June 2024, tutorial. [Online]. Available: \url{https://www.ieee-iv-4dradar.org/}
\BIBentrySTDinterwordspacing

\bibitem{snn_survey}
J.~D. Nunes, M.~Carvalho, D.~Carneiro, and J.~S. Cardoso, ``Spiking neural networks: A survey,'' \emph{IEEE Access}, vol.~10, pp. 60\,738--60\,764, 2022.

\bibitem{direct_snn_survey}
Y.~Guo, X.~Huang, and Z.~Ma, ``Direct learning-based deep spiking neural networks: a review,'' \emph{Frontiers in Neuroscience}, vol.~17, p. 1209795, 2023.

\bibitem{dl_in_snn}
A.~Tavanaei, M.~Ghodrati, S.~R. Kheradpisheh, T.~Masquelier, and A.~Maida, ``Deep learning in spiking neural networks,'' \emph{Neural networks}, vol. 111, pp. 47--63, 2019.

\bibitem{snn_efficiency}
M.~Dampfhoffer, T.~Mesquida, A.~Valentian, and L.~Anghel, ``Are snns really more energy-efficient than anns? an in-depth hardware-aware study,'' \emph{IEEE Transactions on Emerging Topics in Computational Intelligence}, vol.~7, no.~3, pp. 731--741, 2023.

\bibitem{snn_end_to_end}
R.-J. Zhu, Z.~Wang, L.~H. Gilpin, and J.~Eshraghian, ``Autonomous driving with spiking neural networks,'' in \emph{The Thirty-eighth Annual Conference on Neural Information Processing Systems}, 2024.

\bibitem{planning_sota}
S.~Hu, L.~Chen, P.~Wu, H.~Li, J.~Yan, and D.~Tao, ``St-p3: End-to-end vision-based autonomous driving via spatial-temporal feature learning,'' in \emph{European Conference on Computer Vision}.\hskip 1em plus 0.5em minus 0.4em\relax Springer, 2022, pp. 533--549.

\bibitem{lane_snn}
A.~Viale, A.~Marchisio, M.~Martina, G.~Masera, and M.~Shafique, ``Lanesnns: Spiking neural networks for lane detection on the loihi neuromorphic processor,'' in \emph{2022 IEEE/RSJ International Conference on Intelligent Robots and Systems (IROS)}, 2022, pp. 79--86.

\bibitem{da_snn}
G.~Zhuang, Z.~Bing, K.~Huang, and A.~Knoll, ``Toward neuromorphic perception: Spike-driven lane segmentation for autonomous driving using lidar sensor,'' in \emph{2023 IEEE 26th International Conference on Intelligent Transportation Systems (ITSC)}, 2023, pp. 2448--2453.

\bibitem{snn_direct_training}
Y.~Wu, L.~Deng, G.~Li, J.~Zhu, Y.~Xie, and L.~Shi, ``Direct training for spiking neural networks: Faster, larger, better,'' in \emph{Proceedings of the AAAI conference on artificial intelligence}, vol.~33, no.~01, 2019, pp. 1311--1318.

\bibitem{top_down_perception}
C.~D. Gilbert and W.~Li, ``Top-down influences on visual processing,'' \emph{Nature Reviews Neuroscience}, vol.~14, no.~5, pp. 350--363, 2013.

\bibitem{top_down_perception_infant}
G.~D. Reynolds, ``Infant visual attention and object recognition,'' \emph{Behavioral Brain Research}, vol. 285, pp. 34--43, May 15 2015.

\bibitem{lif}
L.~F. Abbott, ``Lapicque's introduction of the integrate-and-fire model neuron (1907),'' \emph{Brain Research Bulletin}, vol.~50, no. 5-6, pp. 303--304, 1999.

\bibitem{snn_first_layer}
P.~U. Diehl and M.~Cook, ``Unsupervised learning of digit recognition using spike-timing-dependent plasticity,'' \emph{Frontiers in Computational Neuroscience}, vol.~9, 2015.

\bibitem{energy_comparison}
M.~Horowitz, ``1.1 computing's energy problem (and what we can do about it),'' in \emph{2014 IEEE International Solid-State Circuits Conference Digest of Technical Papers (ISSCC)}, 2014, pp. 10--14.

\bibitem{kitti}
A.~Geiger, P.~Lenz, C.~Stiller, and R.~Urtasun, ``Vision meets robotics: The kitti dataset,'' \emph{International Journal of Robotics Research (IJRR)}, 2013.

\bibitem{nuscenes}
H.~Caesar, V.~Bankiti, A.~H. Lang, S.~Vora, V.~E. Liong, Q.~Xu, A.~Krishnan, Y.~Pan, G.~Baldan, and O.~Beijbom, ``nuscenes: A multimodal dataset for autonomous driving,'' in \emph{Proceedings of the IEEE/CVF conference on computer vision and pattern recognition}, 2020, pp. 11\,621--11\,631.

\bibitem{astyx}
M.~Meyer and G.~Kuschk, ``Automotive radar dataset for deep learning based 3d object detection,'' in \emph{2019 16th European Radar Conference (EuRAD)}, 2019, pp. 129--132.

\bibitem{vod}
A.~Palffy, E.~Pool, S.~Baratam, J.~F.~P. Kooij, and D.~M. Gavrila, ``Multi-class road user detection with 3+1d radar in the view-of-delft dataset,'' \emph{IEEE Robotics and Automation Letters}, vol.~7, no.~2, pp. 4961--4968, 2022.

\bibitem{tj4d}
L.~Zheng, Z.~Ma, X.~Zhu, B.~Tan, S.~Li, K.~Long, W.~Sun, S.~Chen, L.~Zhang, M.~Wan \emph{et~al.}, ``Tj4dradset: A 4d radar dataset for autonomous driving,'' in \emph{2022 IEEE 25th International Conference on Intelligent Transportation Systems (ITSC)}.\hskip 1em plus 0.5em minus 0.4em\relax IEEE, 2022, pp. 493--498.

\bibitem{rpfa}
B.~Xu, X.~Zhang, L.~Wang, X.~Hu, Z.~Li, S.~Pan, J.~Li, and Y.~Deng, ``Rpfa-net: a 4d radar pillar feature attention network for 3d object detection,'' 09 2021, pp. 3061--3066.

\bibitem{pointpillars}
A.~H. Lang, S.~Vora, H.~Caesar, L.~Zhou, J.~Yang, and O.~Beijbom, ``Pointpillars: Fast encoders for object detection from point clouds,'' in \emph{Proceedings of the IEEE/CVF conference on computer vision and pattern recognition}, 2019, pp. 12\,697--12\,705.

\bibitem{multiframe_4d}
B.~Tan, Z.~Ma, X.~Zhu, S.~Li, L.~Zheng, S.~Chen, L.~Huang, and J.~Bai, ``3-d object detection for multiframe 4-d automotive millimeter-wave radar point cloud,'' \emph{IEEE Sensors Journal}, vol.~23, no.~11, pp. 11\,125--11\,138, 2023.

\bibitem{rcfusion}
L.~Zheng, S.~Li, B.~Tan, L.~Yang, S.~Chen, L.~Huang, J.~Bai, X.~Zhu, and Z.~Ma, ``Rcfusion: Fusing 4-d radar and camera with bird’s-eye view features for 3-d object detection,'' \emph{IEEE Transactions on Instrumentation and Measurement}, vol.~72, pp. 1--14, 2023.

\bibitem{vgg}
K.~Simonyan and A.~Zisserman, ``Very deep convolutional networks for large-scale image recognition,'' \emph{arXiv preprint arXiv:1409.1556}, 2014.

\bibitem{fpn}
T.-Y. Lin, P.~Doll{\'a}r, R.~Girshick, K.~He, B.~Hariharan, and S.~Belongie, ``Feature pyramid networks for object detection,'' in \emph{Proceedings of the IEEE conference on computer vision and pattern recognition}, 2017, pp. 2117--2125.

\bibitem{yolo}
J.~Redmon, ``You only look once: Unified, real-time object detection,'' in \emph{Proceedings of the IEEE conference on computer vision and pattern recognition}, 2016.

\bibitem{spiking_pointnet}
D.~Ren, Z.~Ma, Y.~Chen, W.~Peng, X.~Liu, Y.~Zhang, and Y.~Guo, ``Spiking pointnet: Spiking neural networks for point clouds,'' \emph{Advances in Neural Information Processing Systems}, vol.~36, 2024.

\bibitem{posner}
M.~I. Posner, ``Orienting of attention,'' \emph{Quarterly Journal of Experimental Psychology}, vol.~32, no.~1, pp. 3--25, 1980.

\bibitem{thop}
Lyken17, ``pytorch-opcounter: Count operations in pytorch models,'' \url{https://github.com/Lyken17/pytorch-OpCounter}, 2019.

\end{thebibliography}

\end{document}